%% file: acl.tex
\pdfoutput=1

\documentclass[11pt]{article}

\usepackage{EMNLP2023}
\usepackage{times}
\usepackage{latexsym}

\usepackage[T1]{fontenc}

\usepackage[utf8]{inputenc}

\usepackage{microtype}

\usepackage{inconsolata}

\usepackage{graphicx}
\usepackage{booktabs}
\usepackage{CJKutf8}
\usepackage{makecell}
\usepackage{color}
\usepackage{amsmath}
\usepackage{multirow}
\usepackage{enumitem}
\usepackage{xspace,booktabs}
\newcommand{\our}{\textsc{DDS}\xspace}
\title{Dynamic Stochastic Decoding Strategy for Open-Domain Dialogue Generation}

  \author{Yiwei Li$^1$,  Fei Mi$^2$, Yitong Li$^{2,3}$, Yasheng Wang$^2$, Bin Sun$^1$,Shaoxiong Feng$^1$, Kan Li$^{1}$\footnotemark[1]\\
  { $^1$School of Computer Science \& Technology, Beijing Institute of Technology }\\
  { $^2$ Huawei Noah's Ark Lab \quad  $^3$Huawei Technologies Ltd.} \\
  {\small\texttt{\{liyiwei,binsun,shaoxiongfeng,likan\}@bit.edu.cn} }\\
  {\small \texttt{\{mifei2,liyitong3,wangyasheng\}@huawei.com}}}

\begin{document}
\begin{CJK*}{UTF8}{gbsn}
\maketitle
\renewcommand{\thefootnote}{\fnsymbol{footnote}} 
\footnotetext[1]{Corresponding author.} 
\input{abstract}
\input{introduction}

\input{method}

\input{experiment}
\input{conclusion}


\section*{Limitations}
The following are our limitations:
\begin{itemize}
    \item The contribution for our work may go beyond dialogue generation task. Nowadays, more and more tasks are combined in one model, especially the large language model like ChatGPT. Given that different tasks have different optimal hyper-parameter for decoding temperature, it is badly needed to adjust the temperature adaptively to handle all tasks simultaneously. But we haven't expended proposed strategy to LLMs.
    \item Since there is no suitable public Chinese QA conversational dataset available, the QA datasets we utilize are collected through our internal efforts and haven't been released publicly now. It may be difficult to reproduce our results in this manner.
    \item Considering reranking approach is very popular and effective for text generation, we haven't evaluated the performance of combining it with our proposed method. 
\end{itemize}

\section*{Acknowledgements}
This work is supported by the Beijing Natural Science Foundation, China (Nos. 4222037, L181010).

\bibliography{anthology,custom}

\newpage
\appendix
\input{appendix.tex}

\end{CJK*}

\end{document}

%% file: abstract.tex
\begin{abstract}
Stochastic sampling strategies such as top-$k$ and top-p have been widely used in dialogue generation task.
However, as an open-domain chatting system, there will be two different conversation scenarios, i.e. chit-chat and knowledge-based question answering. In the former situation, responses diversity is essential due to the one-to-many nature in dialogue. The latter, on the other hand, requires less randomness given that stochastic decoding strategy entails the risk of generating incorrect information. As a result, an adaptive and flexible decoding strategy is needed to cope with these two scenarios simultaneously. To this end, we propose the \textbf{d}ynamic \textbf{d}ecoding \textbf{s}trategy (\textbf{\our}), which can adjust the decoding space w.r.t. different contexts. In \our, both sequence-level and token-level adaptive search can be achieved to adjust the decoding process in a unified framework. Besides, our adaptive algorithm can not only be used during model inference, but it can also be applied during the model training stage to further enhance the performance. Comprehensive experiments indicate that the proposed decoding strategy can consistently improve the performance of pre-trained dialogue models when coupled with four well-used stochastic decoding algorithms.
\end{abstract}

%% file: introduction.tex
\section{Introduction}
Building generative open-domain dialogue system is a significant yet challenging area of deep learning research. It has been widely recognized that the pre-training paradigm, in which large-scale transformer-based models are trained with massive amounts of conversational data, is an effective and promising approach. Some of the more notable works in English include DialoGPT \citep{zhang-etal-2020-dialogpt}, LaMDA \citep{lamda},
Blender \citep{roller-etal-2021-recipes,blender3}, 
and lately, ChatGPT has attracted great attention and interest from researchers and the industry. For chinese dialogue models, EVA \citep{eva1,eva2}, PanGu-Bot \citep{pangu} and PLATO \citep{bao-etal-2020-plato,bao-etal-2021-plato,plato-xl} are also excellent options. In recent research, however, it has been demonstrated that decoding strategies play an important role in performance even beyond model architecture \citep{decoding-test}, whereas standard strategies remain relatively unchanged \citep{MBRD}.

\begin{table}[t]
    \centering
    \small
    \begin{tabular}{l}
    \toprule
    \makecell[c]{\textbf{Chit-chat}} \\
    \midrule
         $c$: 我好喜欢猫猫  (I love cats.)\\  
         $r_1$: \textcolor{blue}{我}好\textcolor{blue}{喜欢}我的猫猫啊 \, (\textcolor{blue}{I love} my \textcolor{blue}{cat} so much) \\
         $r_2$:  \textcolor{blue}{我也喜欢猫猫}！ \,(\textcolor{blue}{I like cats, too}!) \\
         $r_3$:  \textcolor{blue}{我也喜欢猫猫} \, (\textcolor{blue}{I like cats, too}) \\
         $r_4$:  \textcolor{blue}{我也喜欢猫猫}！\, (\textcolor{blue}{I like cats, too}!) \\
         $r_5$:  \textcolor{blue}{我也}好\textcolor{blue}{喜欢} \,  (\textcolor{blue}{I like} it \textcolor{blue}{too}) \\
    \toprule
    \makecell[c]{\textbf{Factual Question Answering}} \\
    \midrule
         $c$: 土木工程属于理科还是工科 \, (Does civil engineering \\
         belong to science or engineering)\\  
         $r_1$: 工科,土木是\textcolor{red}{理科} \, (Engineering, civil  \\
         engineering is \textcolor{red}{science}) \\ 
         $r_2$:  土木是工科,\textcolor{red}{土木是建筑} \, (Civil engineering is \\
         engineering, \textcolor{red}{civil engineering is architecture}.) \\
         $r_3$:  工科 \, (Engineering) \\
         $r_4$:  工科 \, (Engineering) \\
         $r_5$:  \textcolor{red}{文科} \, (\textcolor{red}{Liberal arts}) \\
    \bottomrule
    \end{tabular}
    \caption{Generated examples by EVA2.0 on both two scenarios, where top-$k$ sampling
    is used with temperature set to 1. $r_{1-5}$ refer to five generated responses for the same context $c$. Blue part of chit-chat reflects the high similarity of responses, whilst red part reveals the inappropriate answers in factual QA scenario.}
    \label{tb:intro}
    \vspace{-0.4cm}
\end{table}

Stochastic decoding algorithms are widely used for dialogue generation task. 
Users expect varying responses from a chatbot when they input similar queries, or they tend to become bored and lose interest if it only responds with fixed reply. 
For such a chit-chat scenario, deterministic decoding algorithms, such as greedy search or beam search, are not suitable. Additionally, even when using large pre-trained language models, decoding strategies that aim for high probability output, suffer from incredible degeneration issue \citep{top-p,unlikelihood}. Consequently, dialogue generation models are inclined to employ stochastic sampling methods such as top-$k$ sampling \citep{fan-etal-2018-hierarchical} or nucleus sampling \citep{top-p}, where the probability distribution will be shaped by the temperature $T$.

Aside from chit-chat, however, there is another scenario for chatbots, namely factual question answering (QA). 
Unfortunately, 
since the size of the decoding space required for two different dialog scenarios is different, stochastic sampling methods are not able to handle both simultaneously due to the unified and constant randomness of their decoding processes. As shown in Table~\ref{tb:intro}, with the same temperature, the chit-chat sample has a narrow range of generation, where from $r_1$ to $r_5$ are the same \textit{I like cats too}-like responses. Whereas, candidates response to the factual question are too diverse, leading to answers are factually incorrect ($r_1$ and $r_5$), with low fluency ($r_2$) or self-contradictory ($r_1$). As a result, the determined sampling randomness will reduce the diversity under chit-chat condition while enlarge it for question answering, which will increase the risk of generating dull responses and wrong answers.
In addition, even under the same scenario, different contexts will have varying degrees of decoding flexibility \citep{csaky-etal-2019-improving}. For example, \textit{What animals do you like?} has larger response space than \textit{Do you love cats?}. Furthermore, different tokens has different ranges of decoding space within the same utterance \citep{top-p}.

To resolve the drawbacks of existing stochastic decoding algorithms, we propose a dynamic decoding strategy (DDS) for dialogue generation, which can be combined with mainstream stochastic sampling. The key intuition of dynamic sampling is that the decoding space varies according to the context, therefore the shape of probability distribution should be adjusted adaptively. 
To achieve this goal, we incorporate an additional diversity predicting head into the dialogue generation model, which is capable of producing the score based on decoding diversity to guide the sampling process adaptively. It only introduces a few parameters and performs decoding at a similar speed to standard dialogue models. The labeled data for training the head is derived from the pre-trained model automatically. Three types of mapping functions are designed, projecting the diversity score to the temperature for shaping the sampling distribution. In order to control the token generation in a more fine-grained manner, the regression head can be applied to each output token or the whole context, allowing us to control the randomness of decoding at both levels. Apart from inference, adaptive temperature can also be introduced to dialogue training stage to balance the model prediction confidence.

We perform extensive experiments on two union of datasets with two Chinese pre-trained dialogue models. The results show that the DDS can largely improve the performance of four sampling-based decoding algorithms. Human evaluation is also conducted to ensure relevance and fluency of responses while improving diversity.

In summary, our contributions are as follows:
\begin{itemize}[itemsep=1pt,topsep=1pt,parsep=1pt]
    \item We propose a novel dynamic decoding mechanism for dialogue generation, which can easily be integrated into stochastic decoding strategies and handle different conversational scenarios simultaneously.
    \item The mechanism can be conducted on both sentence level and token level with three mapping functions, and adaptive temperature training is introduced except for the inference stage.
    \item Extensive evaluations show that the proposed decoding strategy can largely improve the performance of dialogue models with strong generalization ability when coupled with widely used stochastic decoding strategies. 
\end{itemize}

%% file: method.tex
\section{Background}

\begin{figure*}[t]
\centering
\includegraphics[width=1.0\linewidth]{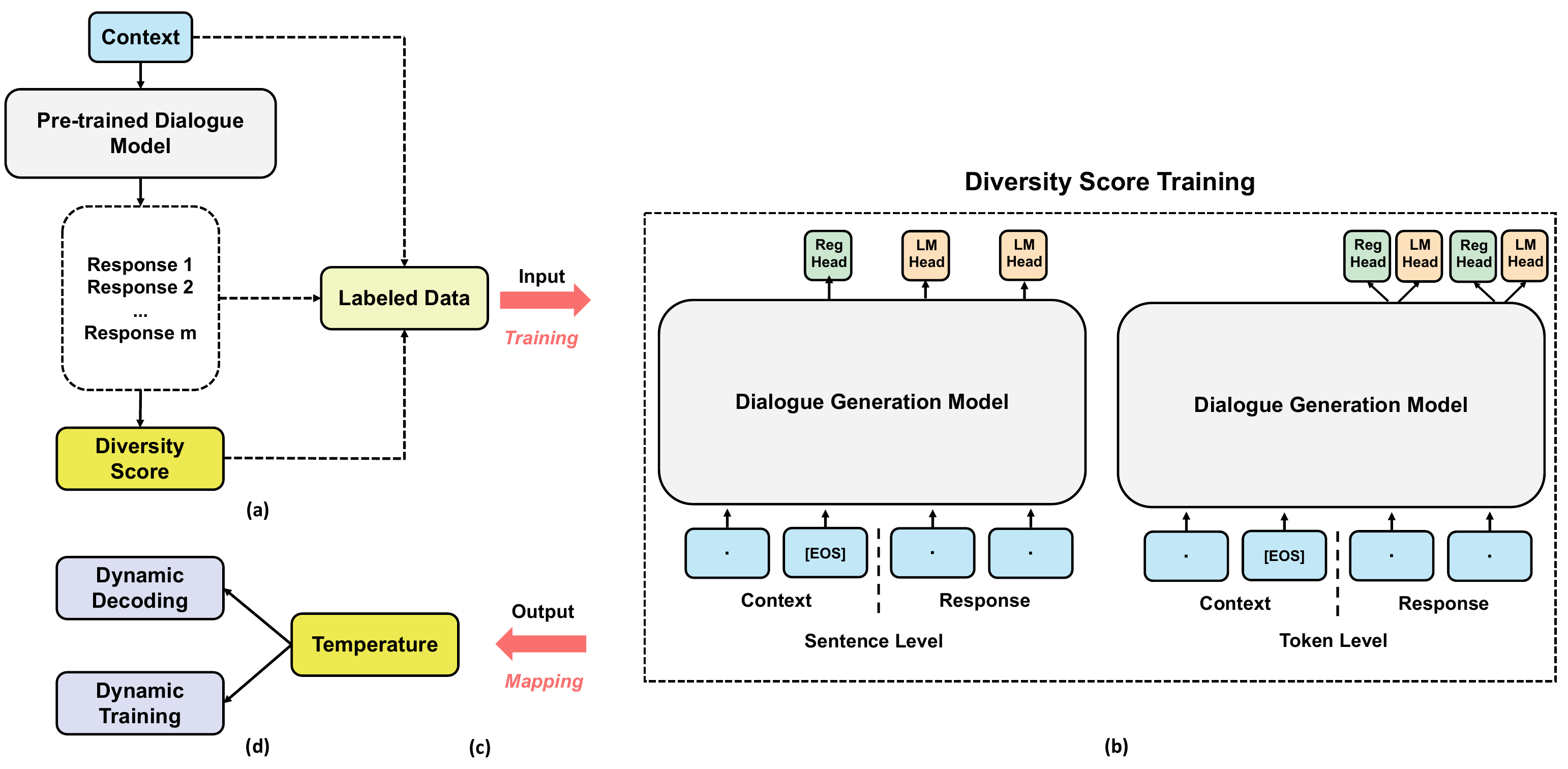}
\caption{An overview of the process of DDS: (a) Calculating the diversity score. (b) Training the regression head. (c) Mapping score to temperature. (d) Dynamic decoding and training.}
\label{fig:method}
\end{figure*}

\subsection{Dialogue Generation}
In this work, we work with the task of dialogue generation in open-domain, where the input context $\boldsymbol{c}=\{c_1, c_2,...\}$ can be either a chat conversation or a factual question and response $\boldsymbol{r}=\{r_1, r_2,...\}$ is produced accordingly. 
Dialogue generation models, which are normally pre-trained on massive conversational corpora nowadays, directly models the response probability $p_\theta(\boldsymbol{r} \mid \boldsymbol{c})$, where $\theta$ indicates the model parameters. 
Standard MLE training is used to minimize the negative log-likelihood (NLL) of the training data:

{
\small
\begin{align}
\nonumber &\mathcal{L}_{\mathrm{NLL}}\left(P_{\text {data }} ; \theta\right)=E_{(\boldsymbol{c}, \boldsymbol{r}) \sim P_{\text {data }}}\left(-\log P_\theta(\boldsymbol{r} \mid \boldsymbol{c})\right) \\
&=E_{(\boldsymbol{c}, \boldsymbol{r}) \sim P_{\text {data }}}(-\sum_{t=1}^T \log P_\theta\left(r_t \mid \boldsymbol{r}_{<t}, \boldsymbol{c}\right)),
\end{align}
}
where $T$ is the length of the response $\boldsymbol{r}$, and the token probability distribution $P_\theta$ is typically modeled as softmax-normalized logits from decoder output $z_t$ by:
\begin{equation}
\small
P_\theta\left(r_t \mid \boldsymbol{r}_{<t}, \boldsymbol{c}\right)=\operatorname{softmax}\left(z_t\right)
\end{equation}

Decoding process is the search for a response token string $\boldsymbol{r^*}$ according to the given dialogue model $\theta$ and context $\boldsymbol{c}$. Most current generative methods employ one of a few standard decoding strategies, which may be characterized as either deterministic or stochastic in nature.

\subsection{Stochastic Decoding Algorithms}
Deterministic decoding algorithms like greedy search or beam search, choose the most probable token or path at each step, generating fixed responses through the following form:
\begin{equation}
\small
\boldsymbol{r}^{\star}=\underset{\boldsymbol{r}}{\operatorname{argmax}} \, p_\theta (\boldsymbol{r} \mid \boldsymbol{c})
\end{equation}

Different from that, stochastic algorithms will generate various responses given the same context by sampling $\boldsymbol{r} \sim p_\theta (\cdot \mid \boldsymbol{c})$. Based on this, four sampling approaches are briefly presented below.

\paragraph{Temperature Sampling.} It is a stochastic sampling method in which the next token is chosen at random based on the new biased probability distribution $p_\theta ^{'}$ shaped by the \textbf{temperature} \textbf{$T$} \citep{temperature}:

\begin{equation}
\small
p^{'}_\theta\left(r_t | \boldsymbol{r}_{<t}, \boldsymbol{c}\right)=\frac{\exp \left(p_\theta\left(r_t | \boldsymbol{r}_{<t}, \boldsymbol{c}\right) / T\right)}{\sum_{r} \exp \left(p_\theta\left(r | \boldsymbol{r}_{<t}, \boldsymbol{c}\right) / T\right)}
\label{eq:tem}
\end{equation}

\paragraph{Top-$k$ Sampling} Based on temperature sampling, it truncates the probability distribution produced by the model by limiting the sampling space to the tokens with top k highest possibilities before sampling \citep{fan-etal-2018-hierarchical}.

\paragraph{Top-p Sampling.} Instead of considering a fixed number of tokens in each decoding step, nucleus (top-p) sampling dynamically selects the smallest set of tokens where the sum of their probabilities is more than the threshold $p$ \citep{top-p}.

\paragraph{Locally Typical Sampling.} It truncates the probability distribution by local informativeness to generate more human-like text \citep{typical}.

\section{Methodology}
We propose the dynamic decoding strategy to dynamically compute temperature $T^{'}$ w.r.t. different contexts, which replaces $T$ in Equation~\ref{eq:tem} for all four sampling methods outlined above. The value of this parameter $T^{'}$ will vary adaptively according to the size of the decoding space. In this section, we first describe how to build the labeled data about dialogue decoding diversity automatically. After that, we elaborate the regression head trained by it for predicting diversity scores on two levels, which will then be projected to temperature $T^{'}$ in accordance with three different mapping strategies. Besides, the dynamic $T^{'}$ can also be applied to training stage. The overview of the proposed framework is illustrated in Figure~\ref{fig:method}.

\subsection{Diversity Score Calculation}
Labeled data is needed to train the regression head to predict the temperature.
$\mathcal{D}=\left\{\left(c_i, r_i\right)\right\}_{i=1}^n$ denotes a training set consisting of $n$ dialogues. In order to quantify the range of decoding space available for a given context $c_i$, we seek to determine its diversity score $s_i$. To achieve this, instead of expensive human annotations, we construct the labeled data automatically. We are motivated by the strong generation capability of pre-trained dialogue models, which has been trained by a large amount of conversational data from various domains. For each $c_i \in \mathcal{D}$, the dialogue model generates m candidates $\{\hat{r_i}\}_m$ based on it, after which the similarity degree between them will be determined. BERTScore \citep{bertscore} is a popular learned evaluation metric for doing this. It compares sentences using contextual embeddings from a pre-trained BERT model, computing a similarity score based on the cosine similarity between the sentence embeddings. We trained the Chinese BERT model on wiki2019zh\footnote{https://github.com/brightmart/nlp\_chinese\_corpus} dataset using the framework from SimCSE \citep{gao-etal-2021-simcse} to calculate the score. The average BERTScore of each $\{\hat{r_i}\}_m$ can reflect the diversity of them, deemed as the range of generation space for the context $c_i$. The higher the score, the narrower the range. Consequently, the labeled dataset $\mathcal{D^{'}}=\left\{\left(c_i, \{\hat{r_i}\}_m, s_i, \right)\right\}_{i=1}^n$ is constructed.

\subsection{Diversity Score Training}
For training and predicting the diversity score efficiently, we design the regression head based on the dialogue generation model, which maps token representation into a one dimensional vector using two feed-forward networks with non-linearity between them:
\begin{equation}
\small
    score = tanh(W_1^Tx + b_1)W_2^T + b_2
\end{equation}
Then, the predicted score $\hat s$ will be fitted to label $s_i$ through MSE loss:
\begin{equation} 
\small
\mathcal{L}_{\mathrm{MSE}}(P_{\text {data }}^{'} ; \theta)=E_{(\boldsymbol{c}, \boldsymbol{s}) \sim P_{\text {data }}^{'}}\left((\boldsymbol{s} - \boldsymbol{\hat s})^{2}\right)
\end{equation}
As shown in Figure~\ref{fig:method}, the regression head can be employed on two levels:
\paragraph{Sentence-level}
On this condition, the diversity score comes from the head of EOS token (denotes the
end of a sentence) of context. Therefore, only $c_i$ and $s_i$ are needed from $\mathcal{D^{'}}$ for training the head. 
\paragraph{Token-level}
For token-level situation, the hidden state of each generated token will provide the diversity score through the regression head. Thus, the head will be trained by each $\hat{r_i} \in \{\hat{r_i}\}_m$ with the same label $s_i$.

There are two ways to train the regression head: either individually with other parameters fixed, or jointly with the standard dialogue generation task. In addition, due to some unexpected samples in $\mathcal{D^{'}}$ (please refer to Table~\ref{tb:intro} and Figure~\ref{fig:score}), the data filtering process will be conducted before training. Afterwards, the predicted diversity score may be more accurate than the one directly derived from the pre-trained model.

\begin{table*}[h]
    \centering
    \small
    \renewcommand\tabcolsep{3.5pt}
    \begin{tabular}{l l c c c c c c c c}
    \toprule
    Datasets & Decoding Strategy & BLEU-1 & BLEU-2	& BLEU-3 & BLEU-4 & F1 & ROUGE-1 & ROUGE-2 & ROUGE-L \\ \toprule
    \multirow{8}{*}{LQA} & Top-$k$ (fixed T) & 0.4327  & 0.2640  & 0.1616  & 0.0988  & 0.2149  & 0.2081  & 0.0412  & 0.1764  \\ 
      & Top-$k$ (\our) & \textbf{0.4410}  & \textbf{0.2701}  & \textbf{0.1659}  & \textbf{0.1019}  & \textbf{0.2187}  & \textbf{0.2083}  & \textbf{0.0452}  & \textbf{0.1827}  \\ \cmidrule(lr){2-10} 
      & Top-p (fixed T)& 0.4109  & 0.2490  & 0.1515  & 0.0924  & 0.1870  & 0.1882  & 0.0325  & 0.1491  \\ 
      & Top-p (\our)  & \textbf{0.4405}  & \textbf{0.2698}  & \textbf{0.1657}  & \textbf{0.1017}  & \textbf{0.2170}  & \textbf{0.2069}  & \textbf{0.0448}  & \textbf{0.1802}  \\ \cmidrule(lr){2-10} 
      & Temperature (fixed T) & 0.3891  & 0.2342  & 0.1416  & 0.0856  & 0.1679  & 0.1710  & 0.0254  & 0.1337  \\ 
      & Temperature (\our)& \textbf{0.4357}  & \textbf{0.2663}  &\textbf{ 0.1633}  & \textbf{0.1001}  & \textbf{0.2128 } & \textbf{0.2062}  & \textbf{0.0427}  & \textbf{0.1745} \\
      \cmidrule(lr){2-10} 
      & Typical (fixed T) & 0.3971          & 0.2393          & 0.1447          & 0.0876          & 0.1770          & 0.1777          & 0.0263          & 0.1392          \\
      & Typical (\our)& \textbf{0.4378} & \textbf{0.2682} & \textbf{0.1649} & \textbf{0.1014} & \textbf{0.2169} & \textbf{0.2073} & \textbf{0.0451} & \textbf{0.1791} \\  \midrule
     \multirow{8}{*}{PersonQA} & Top-$k$ (fixed T) &  0.5751  & 0.4618  & 0.3840  & 0.3258  & 0.4321  & 0.4234  & 0.3203  & 0.4284  \\
      &Top-$k$ (\our) &  \textbf{0.6137}  & \textbf{0.4989}  & \textbf{0.4200}  & \textbf{0.3609}  & \textbf{0.4619}  & \textbf{0.4524}  & \textbf{0.3533}  & \textbf{0.4590 } \\ \cmidrule(lr){2-10} 
      &Top-p (fixed T)&0.5400  & 0.4358  & 0.3647  & 0.3117  & 0.4044  & 0.3962  & 0.3041  & 0.4010  \\ 
      &Top-p (\our)  & \textbf{0.5979}  & \textbf{0.4874}  & \textbf{0.4114}  & \textbf{0.3539}  & \textbf{0.4488 } & \textbf{0.4403}  & \textbf{0.3461}  & \textbf{0.4456}  \\ \cmidrule(lr){2-10} 
      &Temperature (fixed T) & 0.5413  & 0.4365  & 0.3647  & 0.3112  & 0.4038  & 0.3958  & 0.3024  & 0.4008  \\ 
      &Temperature (\our)& \textbf{0.5894}  & \textbf{0.4811}  & \textbf{0.4066}  & \textbf{0.3506}  & \textbf{0.4439}  & \textbf{0.4346}  & \textbf{0.3417}  & \textbf{0.4407}  \\
      \cmidrule(lr){2-10} 
      &Typical (fixed T) & 0.5348  & 0.4317  & 0.3611  & 0.3082  & 0.3994  & 0.3916  & 0.3010  & 0.3962  \\
      &Typical (\our)& \textbf{0.5963}  & \textbf{0.4872}  & \textbf{0.4121}  & \textbf{0.3555}  & \textbf{0.4495}  & \textbf{0.4407}  & \textbf{0.3477}  & \textbf{0.4469} \\ 
        \bottomrule
    \end{tabular}
        \begin{tabular}{l l c c c c c c c}
    \toprule
      Datasets &  Decoding Strategy & Distinct-1 & Distinct-2	& Distinct-3 & Ent-1 & Ent-2 & Ent-3 & BERTScore\\  \toprule
        \multirow{8}{*}{LCCC} & Top-$k$ (fixed T) & 0.1015  & 0.3973  & 0.6659  & 10.0321  & 18.5411  & 19.6180 &0.5764  \\
        &Top-$k$ (\our) & \textbf{0.1036}  & \textbf{0.4119}  & \textbf{0.6889}  & \textbf{10.0755}  & \textbf{18.6606}  & \textbf{19.8775} &\textbf{0.5617} \\ \cmidrule(lr){2-9} 
        &Top-p (fixed T) & 0.1523  & 0.6170  & 0.9057  & 11.1319  & 18.9154  & 20.4290 &0.4562 \\
            &Top-p (\our) & \textbf{0.2101}  & \textbf{0.7718}  & \textbf{0.9428}  & \textbf{12.5948 } & \textbf{19.4330}  & \textbf{21.4829} &\textbf{0.4332}  \\ \cmidrule(lr){2-9} 
        &Temperature (fixed T) & 0.1818  & 0.6866  & 0.9418  & 11.6779  & 19.0928  & 20.7616 &0.4424 \\
        &Temperature (\our) & \textbf{0.2555 } & \textbf{0.8685}  & \textbf{0.9867}  & \textbf{13.1907}  & \textbf{19.5489}  & \textbf{21.7447} &\textbf{0.4243} \\ \cmidrule(lr){2-9} 
        &Typical (fixed T) & 0.1519  & 0.6132  & 0.8929  & 11.1861  & 18.9442  & 20.4646 &0.4578 \\ 
        &Typical (\our) &  \textbf{0.2331}  & \textbf{0.8133}  & \textbf{0.9646}  & \textbf{12.6864}  & \textbf{19.4573}  & \textbf{21.4895} &\textbf{0.4321} \\ \midrule
        \multirow{8}{*}{Diamante} & Top-$k$ (fixed T) & 0.1124  & 0.4100  & 0.6502  & \textbf{10.1575}  & 12.4041  & 15.6205  &0.6532  \\ 
        &Top-$k$ (\our) &\textbf{0.1153}  & \textbf{0.4175}  & \textbf{0.6628}  & 10.1398  & \textbf{12.4172}  & \textbf{15.6745}&\textbf{0.6438} \\ \cmidrule(lr){2-9} 
        &Top-p (fixed T) &  0.1282  & 0.4582  & 0.7036  & 10.2857  & 12.6627  & 15.8346  &0.6144  \\
        &Top-p (\our) & \textbf{0.1791}  & \textbf{0.5401}  & \textbf{0.7811}  & \textbf{10.4122}  & \textbf{12.9131}  & \textbf{16.0630}  &\textbf{0.5822} \\ \cmidrule(lr){2-9} 
        &Temperature (fixed T) &  0.1408  & 0.5098  & 0.7744  & 10.3355  & 12.7581  & 15.9274 &0.4591  \\
        &Temperature (\our) & \textbf{0.2377}  & \textbf{0.6362}  & \textbf{0.8510}  & \textbf{10.5948}  & \textbf{13.2204}  & \textbf{16.2846 } &\textbf{0.4339} \\ \cmidrule(lr){2-9} 
        &Typical (fixed T) &  0.1267  & 0.4545  & 0.7038  & 10.3077  & 12.6324  & 15.7905 &0.4627  \\
        &Typical (\our) &\textbf{0.2601}  & \textbf{0.6172}  & \textbf{0.8237}  & \textbf{10.4760}  & \textbf{12.9582 } & \textbf{16.0774} &\textbf{0.4234} \\
        \bottomrule
    \end{tabular}
    \caption{Automatic evaluations results on PanGu-Bot. DDS has significantly improved the performance of all four well-known stochastic decoding algorithms on four datasets.}
    \label{tb:pangu}
\end{table*}

\subsection{Temperature Mapping Strategies}
\begin{figure}[t]
\centering
\includegraphics[width=0.85\linewidth]{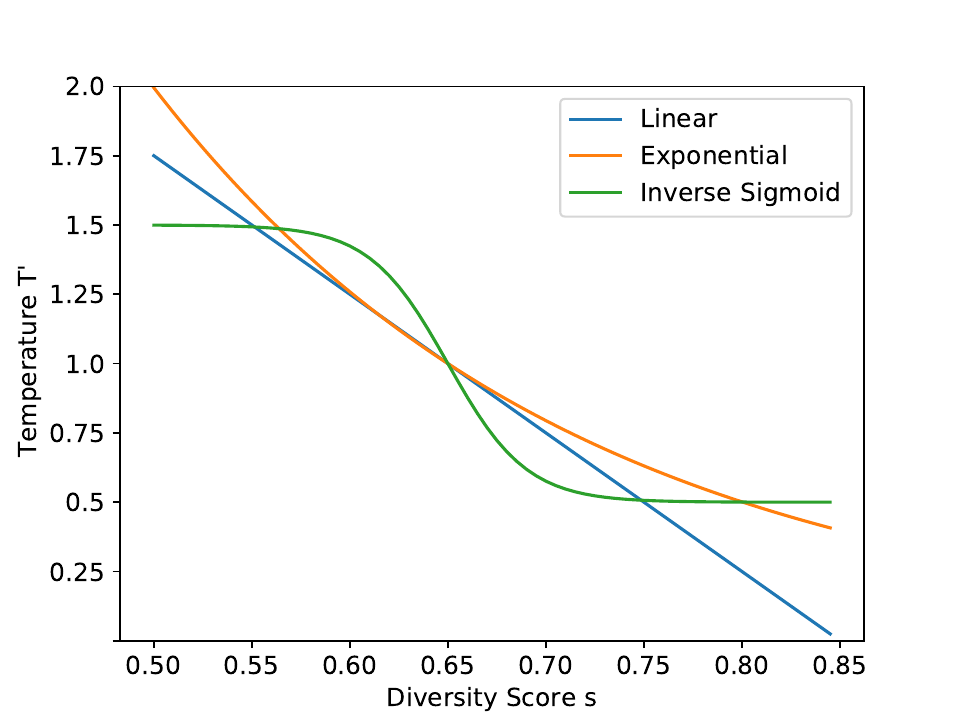}
\caption{Different mapping strategies to project the diversity score to temperature.}
\label{fig:mapping}
\end{figure} 
After obtaining the diversity score $s_i$, we further convert it to guide the dynamic temperature $T^{'}$ for Equation~\ref{eq:tem}.
As $s_i$ increases, $T^{'}$ should decrease to sharpen the probability distribution of sampling and vice versa.
Consequently, three mapping strategies are designed:

\begin{itemize}[align = left, wide=1pt, itemsep=1pt, parsep=1pt, topsep=1pt]
    \item Linear Mapping
    \begin{equation}
    \small
       T(s) = hs + t_0,
    \end{equation}
    where k is the slope.
    \item Exponential Mapping
    \begin{equation}
    \small
    T(s) = h^s + t_0,
    \end{equation}
    where $h < 1$ is the radix to adjust the sharpness of mapping function.
    \item Inverse Sigmoid Mapping
    \begin{equation}
    \small
    T(s)=\frac{h}{h+e^{\frac{s}{h}}} + t_0, 
\end{equation}
where $\mathrm{e}$ is the mathematical constant, and $h \leq 1$ is a hyperparameter to adjust the sharpness. All $t_0$ is the offset to make $T(s)$ equals 1 when $s$ reaches the mean value.
\end{itemize}
A visual representation of different mapping strategies is provided in Figure~\ref{fig:mapping}. In this way, a dynamic temperature $T^{'}$ can be constructed to guide the decoding process adaptively.

\subsection{Dynamic Temperature in Training}
In addition, same as the inference stage, the temperature $T^{'}$ can shape the probability distribution $p_\theta$ of decoder output $z$ during training process by: 
\begin{equation}
\small
p^{i}_\theta=\frac{\exp (z_{i} / T^{'})}{\sum_{j} \exp \left(z_{j} / T^{'}\right)},
\end{equation}
Thus, the dynamic temperature training can be conducted to balance the model prediction confidence of chit-chat and factual question answering scenarios respectively. Considering the one-to-many labels, the former is suitable for low confidence training, whereas the latter requires a higher degree of confidence due to the certainty of the knowledge.


%% file: experiment.tex
\section{Experiments}

\begin{table*}[th]
    \centering
    \small
    \renewcommand\tabcolsep{3.5pt}
    \begin{tabular}{l c c c c c c c c}
    \toprule
        Decoding Strategy & BLEU-1 & BLEU-2	& BLEU-3 & BLEU-4 & F1 & ROUGE-1 & ROUGE-2 & ROUGE-L \\ \toprule
      Top-$k$ (fixed T) &  0.0823  & 0.0495  & 0.0299  & 0.0180  & 0.1139  & 0.0983  & 0.0113  & 0.0988  \\
      Top-$k$ (\our) & \textbf{0.0921}  & \textbf{0.0557}  & \textbf{0.0339}  & \textbf{0.0206}  & \textbf{0.1181}  & \textbf{0.1010}  & \textbf{0.0136}  & \textbf{0.1043}  \\ \midrule
      Top-p (fixed T)&0.0844  & 0.0509  & 0.0309  & 0.0187  & 0.1143  & 0.0990  & 0.0115  & 0.0984  \\
      Top-p (\our)  & \textbf{0.0927}  & \textbf{0.0558}  & \textbf{0.0337}  & \textbf{0.0203}  & \textbf{0.1172 } & \textbf{0.1006}  & \textbf{0.0130}  & \textbf{0.1024}  \\ \midrule
      Temperature (fixed T)&  0.0762  & 0.0452  & 0.0271  & 0.0162  & 0.0656  & 0.0586  & 0.0028  & 0.0568  \\ 
      Temperature (\our)& \textbf{0.0801}  & \textbf{0.0482}  & \textbf{0.0292}  & \textbf{0.0177}  & \textbf{0.1041}  & \textbf{0.0896}  & \textbf{0.0115}  & \textbf{0.0918} \\\midrule
      Typical (fixed T) & 0.0554  & 0.0331  & 0.0200  & 0.0120  & 0.0853  & 0.0724  & 0.0049  & 0.0743  \\
      Typical (\our)& \textbf{0.0923}  & \textbf{0.0555}  & \textbf{0.0336} & \textbf{0.0202}& \textbf{0.1106}  & \textbf{0.0931}  & \textbf{0.0116}  & \textbf{0.0958} \\
        \bottomrule
    \end{tabular}
        \begin{tabular}{l c c c c c c c c}
    \toprule
        Decoding Strategy & Distinct-1 & Distinct-2	& Distinct-3 & Ent-1 & Ent-2 & Ent-3 & BERTScore\\  \toprule
        Top-$k$ (fixed T) & 0.1616  & 0.4769  & 0.7140  & 9.8991  & 18.5029  & 19.3731 &0.6435  \\
        Top-$k$ (\our) & \textbf{0.1639}  & \textbf{0.4950}  & \textbf{0.7510}  & \textbf{9.9633}  & \textbf{18.5990}  & \textbf{19.5902}  &\textbf{0.6320} \\ \midrule
        Top-p (fixed T) & \textbf{0.2055}  & 0.6806  & 0.9368  & 10.4591  & 18.6561  & 19.8080 &0.4890 \\
        Top-p (\our) & 0.2041  & \textbf{0.7127}  & \textbf{0.9490}  & \textbf{10.7369}  & \textbf{18.9950}  & \textbf{20.4575} &\textbf{0.4645} \\ \midrule
        Temperature (fixed T) & 0.3281  & 0.8505  & 0.9841  & 11.9063  & 19.1125  & 20.7625 &0.4213  \\
        Temperature (\our) & \textbf{0.4408}  & \textbf{0.9693}  & \textbf{0.9991}  & \textbf{14.3922}  & \textbf{19.6696}  & \textbf{22.0616} &\textbf{0.4078} \\ \midrule
        Typical (fixed T) &  \textbf{0.1884}  & 0.6393  & 0.9152  & 10.2947  & 18.7910  & 20.0862 &0.4657 \\ 
        Typical (\our) & 0.1708  & \textbf{0.6423}  & \textbf{0.9270}  & \textbf{10.6164}  & \textbf{19.3449}  & \textbf{21.3205} &\textbf{0.4536} \\
        \bottomrule
    \end{tabular}
    \caption{Zero-shot automatic evaluations results of LQA (Up) and LCCC (Down) on EVA2.0.}
    \label{tb:eva}
\end{table*}

\subsection{Dataset}
\begin{figure}[th]
\centering
\includegraphics[width=1.0\linewidth]{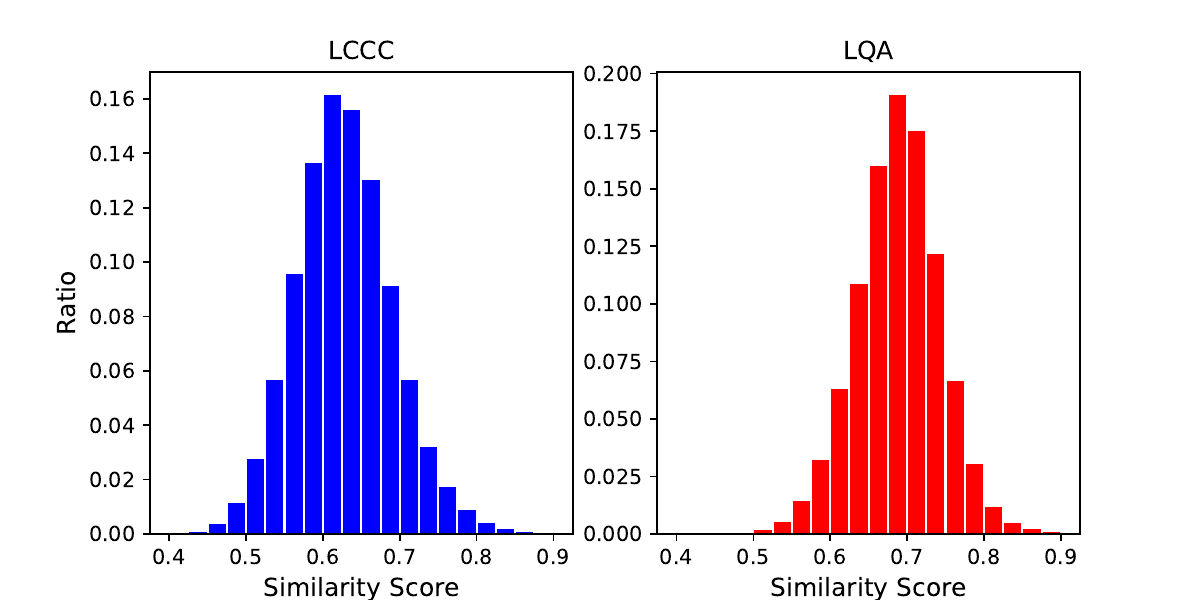}
\caption{Similarity score distributions of LCCC (left) and LQA (right). The former is a chit-chat dataset and the latter is for QA scenario. The samples are generated by PanGu-Bot and the scores are calculated by BERTScore. Although overall scores of the chatting scene are lower, there are also some noise samples with much higher similarity scores for chitchat and lower scores for QA.}
\label{fig:score}
\end{figure}

\begin{table}[t]
\centering
\small
\renewcommand\tabcolsep{8.5pt}
\begin{tabular}{llccc}
\toprule
    & Datasets        & \# Train & \# Valid & \# Test \\ \midrule
 \multirow{2}{*}{$U_S$}   & PersonQA         & 4500 & 500 & 919 \\
    & Diamante        & 3000 & 500 & 916 \\ \midrule
 \multirow{2}{*}{$U_L$}   & LQA         & 115k & 10k & 10k \\
    & LCCC        & 90k & 10k & 10k \\
\bottomrule
\end{tabular}
\caption{Data statistics of the experiment corpora.}
\label{tb:data_statistics}
\end{table}
For training, we use two datasets with different data size to verify the effectiveness of the proposed decoding strategy in two conversation scenarios, each of which contains a chit-chat and a QA dataset.
The first is the union ($U_S$) of Diamante \citep{Diamante}, a human-written chit-chat dialogue dataset, and PersonQA, a question answering data about persons. Both of them are small but with high-quality.
The second dataset ($U_L$) has much larger size, consisting of LCCC-base \citep{LCCC}, and LQA, which includes longer explanations in responses.
We calculate the diversity score of each dataset, and then mix the data within the same union.
Figure~\ref{fig:score} depicts the similarity scores of LCCC and LQA, showing that QA scenario scores are holistically larger than those of chit-chat. 
The overall trend is in line with expectations, while there are some noise samples with much higher scores in LCCC and lower ones in LQA. Table~\ref{tb:intro} shows the cases from those parts and it is what we need to solve through our method.
Therefore, we filter these extreme data by dropping samples whose score is lower than 0.6 in QA dataset and higher than 0.7 in chit-chat dataset. 
Table~\ref{tb:data_statistics} provides the statistics of both unions for training the regression head. Please see Appendix~\ref{QA_data} for more details about QA dataset.
For test, all the four sub-sets are evaluated separately. In this work, we mainly focus on Chinese datasets, but we also conduct additional test in Section~\ref{multi} to verify the multilingual availability.

\label{calculation}

\subsection{Training Settings}
We take two Chinese pre-trained models: PanGu-Bot \citep{pangu} containing 350M parameters and EVA2.0 \citep{eva2} with 300M parameters as the underlying generation models to demonstrate that our method is applicable to a wide range of architectures.
The regression head is trained for 3 epochs and only takes 0.27\% and 0.20\% parameters for PanGu-Bot and EVA2.0 respectively.
\our is introduced to four widely used stochastic decoding strategies at sentence level with inverse sigmoid mapping. 
We set $k=3$, $p=0.9$, $\tau=0.9$ for top-$k$, top-p, typical sampling respectively, and $T=1$ for all of them including Temperature sampling as common settings. In main experiments, we adopt sentence-level DDS, given that its lower costs than token-level one.
The responses are generated 5 times per test.

\subsection{Automatic Evaluation}
For automatic evaluation, we divide metrics into two groups because chit-chat and QA datasets require different evaluation aspects. For factual QA datasets, the most important thing is to verify the knowledge accuracy w.r.t. the ground truth, thus we adopt the following metrics:
\textbf{BLEU-\{1,2,3,4\}} \citep{chen-cherry-2014-systematic},
\textbf{Rouge-\{1,2,L\}} \citep{lin-2004-rouge} and \textbf{F1}.
While for chatting datasets, considering there will be multiple responses for one context, the metrics above are not suitable. Therefore, we utilize these three metrics to evaluate the diversity:
\textbf{Distint-\{1,2,3\}} \citep{li-etal-2016-diversity}, \textbf{Ent-\{1,2,3\}} (word entropy) \citep{csaky-etal-2019-improving} and \textbf{BERTScore} (calculating the similarity score between five generated responses given the same context).

Table~\ref{tb:pangu} shows the results from PanGu-Bot. As can be seen, the proposed dynamic decoding strategy (DDS) improves the performance of all four well-known stochastic decoding algorithms on four datasets, confirming its general applicability and superiority. 
Specifically, for LQA and PersonQA, all metrics obtains the best scores, indicating that DDS can generate more accurate answers for QA scenario. Under the same settings, the higher Distinct and Ent scores of Diamante and LCCC verify the diversity in chit-chat scenario. Appendix~\ref{case} shows some generated cases.
Table~\ref{tb:eva} summarizes the result from EVA2.0 in a zero-shot setting, which illustrates similar trends. This observation demonstrates that the proposed \our can be applied to different model architectures and learning manners.

 \begin{table}[t]
    \centering
    \small
    \begin{tabular}{l c c c}
    \toprule
        Decoding Strategy &  Flu. (\%)& Rel. (\%) & Kappa  \\  \midrule
        Top-$k$ (fixed T) & 97.6 & 59.0 & 0.618 \\
        Top-$k$ (\our)  & 98.3 & 70.0 & 0.439 \\
        \midrule
        Top-p (fixed T) & 92.0 & 62.3 & 0.734\\
        Top-p (\our)  & 90.3 & 60.3 &  0.655\\
        \midrule
        Temperature (fixed T) & 80.3 & 52.7 & 0.496 \\
        Temperature (\our) & 79.0 & 50.0 & 0.512 \\
        \midrule
        Typical (fixed T) & 84.0 & 54.7 & 0.621\\
        Typical (\our)  & 87.3 & 54.7 & 0.431\\
        \bottomrule
    \end{tabular}
    \caption{Human evaluations results on Diamante.}
    \label{tb:human}
\end{table}

\subsection{Human Evaluation}
For chit-chat dataset, although label-related metrics are not suitable, it is also necessary to evaluate its relevance (\textbf{Rel.}) and fluency (\textbf{Flu.}) besides the diversity. So we conduct human evaluation as a supplement to automatic experiment.
\textbf{Rel.} reflects how likely the generated response is relevant to its context. \textbf{Flu.} reflects how likely the generated response comes from human. We collect 100 samples for each decoding setting from Diamante and employ three annotators to judge whether the response is in compliance with above standards. Table~\ref{tb:human} summarizes the human evaluation results. We can see that the proposed approach has similar results compared with baselines, which indicates that dynamic decoding method maintains the relevance and fluency of responses while improving its diversity. 
We use Fleiss's kappa \citep{fleisskappa/measuring} to measure the inter-annotator agreement.

\subsection{Multilingual Availability}
\label{multi}
\begin{table}[th]
    \centering
    \small
    \renewcommand\tabcolsep{4.0pt}
    \begin{tabular}{l c c c c}
    \toprule
       \textit{CQ} & BLEU-4 & F1  & ROUGE-2 & ROUGE-L \\ \midrule
      Base   & 0.0520  & 0.0759   & 0.0133  & 0.0741 \\ 
      \our & \textbf{0.0532}  & \textbf{0.0793}  & \textbf{0.0142}  & \textbf{0.0722}  \\ \midrule
        Base   & 0.0674  & 0.1105   & 0.0391  & 0.1072 \\ 
      \our & \textbf{0.0691}  & \textbf{0.1154}  & \textbf{0.0406}  & \textbf{0.1115}  \\ 
        \toprule
      \textit{Daily} & Dist-2	& Dist-3 & Ent-2 & Ent-3  \\ \midrule
      Base  & 0.2647  & 0.4371  & 14.2122  & 17.5430 \\ 
      \our   & \textbf{0.4023}  & \textbf{0.6056}   & \textbf{14.5874}  & \textbf{17.6932} \\ \midrule
      Base  & 0.2966  & 0.4722  & 13.6437  & 17.2051 \\ 
      \our   & \textbf{0.4158}  & \textbf{0.6141}   & \textbf{13.8967}  & \textbf{17.3642} \\ 
        \bottomrule
    \end{tabular}
    \caption{Zero-shot results on Llama-2-7b \citep{llama2} (Up) and GPT-3.5-turbo (Down). Base means sampling with fixed temperature. \textit{CQ} refers to ComplexQuestions and \textit{Daily} refers to DailyDialog.}
    \label{tb:multi}
\end{table}

Although the proposed method was tested on Chinese corpora, it could work for other languages as well. To demonstrate this, we select English datasets as additional study, ComplexQuestions \citep{CQ} for QA and DailyDialog \citep{dailydialog2017} for chit-chat. The superior results from Table~\ref{tb:multi} with top-p sampling support the multilingual availability of DDS. The linguistic phenomena in English differ greatly from those in Chinese, making this experiment a good test of the applicability of the proposed method to non-Chinese languages.

\subsection{Token Level DDS}

\begin{figure}[thb!]
\centering
\includegraphics[width=0.75\linewidth]{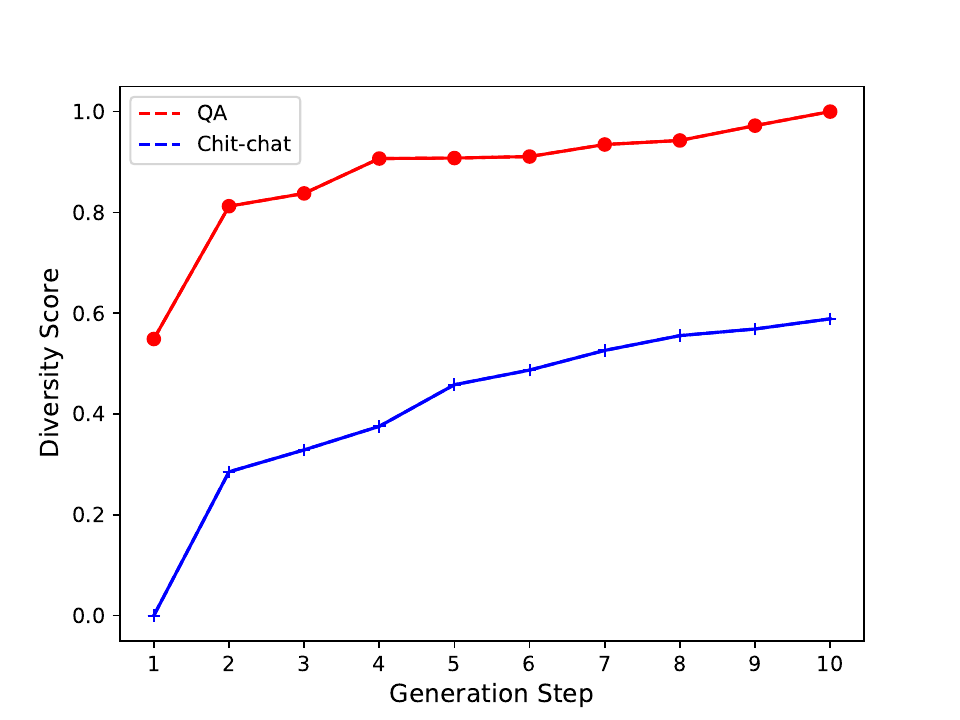}
\caption{Token level diversity score (normalized) over generation steps.}
\label{fig:token}
\end{figure} 

\begin{table}[t]
    \centering
    \small
    \renewcommand\tabcolsep{4pt}
    \begin{tabular}{l c c c c}
    \toprule
       \textit{PersonQA}  & BLEU-4 & F1  & ROUGE-2 & ROUGE-L \\ \midrule
      Base   & 0.3117 & 	0.4044 	 & 	0.3041&  	0.4010   \\ 
      Sent & \textbf{0.3539} &	\textbf{0.4488} 	 &	\textbf{0.3461}& \textbf{0.4456}   \\ 
      Token   & 0.3357 &	0.4335 	&	0.3273 &	0.4303  \\ 
        \toprule
       \textit{Diamante} & Dist-2	& Dist-3 & Ent-2 & Ent-3  \\ \midrule
      Base  & 0.4582 &	0.7036 & 	12.6627 &	15.8346  \\ 
      Sent   & 0.5401 &	0.7811 &	12.9131 &	16.0630 \\ 
      Token   & \textbf{0.5603} &	\textbf{0.8289} 	& 	\textbf{13.1880} &	\textbf{16.2892}   \\ 
        \bottomrule
    \end{tabular}
    \caption{Results of token-level DDS with top-p sampling.}
    \label{tb:token}
\end{table}
Dynamic decoding at the token level is more fine-grained than that at the sentence level. The Figure~\ref{fig:token} depicts that the diversity score (the higher, the narrower decoding space) shows a rising trend over the generation step, which is consistent with the heuristic motivation of \citet{fact} that generating the latter part of a sentence require less decoding randomness. Table~\ref{tb:token} shows the results at both two levels. The scores of token level on both two datasets are higher than base, verifying the effectiveness of it. Different from Diamante, PersonQA does not perform better at the token level than it does at the sentence level. This may be because the higher randomness of former part within the utterance than sentence level, thus it needs further design for mapping strategy. Figure~\ref{fig:token} has shown the effectiveness of predicting diversity score at token level, and we leave the study of exploiting the potential of it as future work.

\subsection{Study of mapping strategies}
\label{sc:map}

\begin{table}[t]
    \centering
    \small
    \renewcommand\tabcolsep{4.5pt}
    \begin{tabular}{l c c c c}
    \toprule
       Mapping & BLEU-4 & F1  & ROUGE-2 & ROUGE-L \\ \midrule
      Identity   & 0.0924  & 0.1870   & 0.0325  & 0.1491  \\ 
      Linear & 0.1004  & 0.2124  & 0.0441  & 0.1753  \\ 
      Exp   & 0.1001  & 0.2100    & 0.0427  & 0.1719 \\ 
      Sigmoid   & 0.1017  & 0.2170   & 0.0448  & 0.1802 \\ 
        \toprule
       Mapping & Dist-2	& Dist-3 & Ent-2 & Ent-3  \\ \midrule
      Identity  & 0.6170  & 0.9057  & 18.9154  & 20.4290 \\ 
      Linear   & 0.7491  & 0.9600   & 19.2278  & 21.0573 \\ 
      Exp   & 0.7760  & 0.9406    & 19.2988  & 21.2100  \\ 
      Sigmoid    & 0.7718  & 0.9428    & 19.4330  & 21.4829  \\ 
        \bottomrule
    \end{tabular}
    \caption{Study of mapping strategies with top-p sampling on LQA (Up) and LCCC (Down).}
    \label{tb:mapping}
\end{table}

\begin{table}[t]
    \centering
    \small
    \begin{tabular}{l c c c c}
    \toprule
       Slope & BLEU-4 & F1  & ROUGE-2 & ROUGE-L \\ \midrule
              Base & 0.0924  & 0.1870   & 0.0325  & 0.1491  \\ 
               1  & 0.0933  & 0.1903    & 0.0345  & 0.1532 \\ 
               2 & 0.0963  & 0.1993   & 0.0378  & 0.1607  \\ 
               3  & 0.0977  & 0.2021    & 0.0387  & 0.1637  \\ 
               4 & 0.0995  & 0.2077   & 0.0419  & 0.1696  \\
               5 & 0.1004  & 0.2124  & 0.0441  & 0.1753  \\ 
        \bottomrule
    \end{tabular}
    \caption{Study of the value of slope.}
    \label{tb:slope}
\end{table}

In this section, we study the effectiveness of different mapping strategies. As shown in Table~\ref{tb:mapping}, all three types of mapping functions can largely improve the performance on both two scenarios. We simply set h for them as 5, 0.01 and 0.02 respectively and actually the hyperparameters do not need to be specially adjusted. For example, the slope of linear mapping can influence the performance, but as shown in Table~\ref{tb:slope}, all five different values can outperform the fixed temperature sampling.

\subsection{Domain Adaptation}

\begin{table}[t]
    \centering
    \small
    \renewcommand\tabcolsep{4.0pt}
    \begin{tabular}{l c c c c}
    \toprule
       \textit{BaikeQA} & BLEU-4 & F1  & ROUGE-2 & ROUGE-L \\ \midrule
      Base   & 0.0924  & 0.1870   & 0.0325  & 0.1491  \\ 
      \our & \textbf{0.1004}  & \textbf{0.2124}  & \textbf{0.0441}  & \textbf{0.1753}  \\ 
        \toprule
       \textit{CDConv} & Dist-2	& Dist-3 & Ent-2 & Ent-3  \\ \midrule
      Base  & 0.6170  & 0.9057  & 18.9154  & 20.4290 \\ 
      \our   & \textbf{0.7491}  & \textbf{0.9600}   & \textbf{19.2278}  & \textbf{21.0573} \\ 
        \bottomrule
    \end{tabular}
    \caption{Results of out-of-domain test.}
    \label{tb:domain}
\end{table}

We conduct experiments with out-of-domain test data on EVA2.0 for further generalization evaluation.
For chit-chat scenario, we choose CDConv \citep{CDConv}, a high-quality dataset for detecting contradiction problem. We only select the first turn of each conversations, where the query is basically the question in chit-chat scenario. For QA scenario, we employ BaikeQA, a QA dataset from Chinese Wiki. The results from Table~\ref{tb:domain} show that \our can still outperform the basic decoding strategy, which indicates the generalization ability.

\subsection{Dynamic Training}

\begin{table}[t]
    \centering
    \small
    \renewcommand\tabcolsep{4.5pt}
    \begin{tabular}{lcccc}
    \toprule
       \textit{PersonQA} & BLEU-4 & F1  & ROUGE-2 & ROUGE-L \\ \midrule
      Base & 0.3117 &	0.4044 	 &	0.3041 &	0.4010    \\ 
      DT &  0.3838 &	0.4758 &	0.3776 &	0.4737  \\ 
      DT+DDS &  \textbf{0.4050} &	\textbf{0.4956} 	& 	\textbf{0.3967} &	\textbf{0.4936} \\ 
        \toprule
       \textit{Diamante} & Dist-2 & Dist-3 & Ent-2 & Ent-3  \\ \midrule
      Base &  0.4582 &	0.7036&  12.6627 &	15.8346  \\ 
      DT &  0.4794 &	0.7428& 12.7369 &	15.9257  \\ 
      DT+DDS & \textbf{0.5479} &	\textbf{0.7986} & \textbf{13.1270} &	\textbf{16.2207}   \\ 
        \bottomrule
    \end{tabular}
    \caption{Results of DT with top-p sampling.}
    \label{tb:dt}
\end{table}

To evaluate the effectiveness of dynamic training (DT), we train the LM head and regression head jointly. The results of Table~\ref{tb:dt} show that dynamic training is effective in improving performance. The dynamic training and decoding can be performed simultaneously, and the higher performance of DT+DDS indicates that the performance can be further enhanced. 

%% file: conclusion.tex
\section{Conclusion}
In this paper, we discuss the drawbacks of commonly used standard decoding methods for open-domain dialogue generation task.
To overcome them, we present a novel dynamic decoding strategy, \our, to handle different conversational scenarios concurrently. It can adaptively adjust the decoding space according to different contexts at both sequence and token levels with three mapping functions.
Moreover, we further boost the performance by introducing the dynamic temperature to training stage.
Extensive experiments demonstrate the superiority and generalization of proposed decoding method.

%% file: appendix.tex


\section{Case Study}
\label{case}
Table~\ref{tb:my_label} and ~\ref{tb:my_label2} shows some cases generated 5 times by the proposed method and baselines on both two scenarios. From where we can see that the diversity of responses under chit-chat is enhanced and the correctness in QA scenario is also improved.
\begin{table}[!ht]
    \centering
    \small
    \begin{tabular}{r|p{5.5cm}}
    \toprule
      \textit{context:}    &  马上就要填报志愿了，你打算学什么专业啊 (It's time to fill in your application. What major do you plan to study)\\
      \midrule
       \multirow{5}{*}{DDS:} & 我要学医,你呢? (I want to study Medicine. How about you?)\\
        & 我想报法学专业,你有什么好的建议吗? (I want to apply for Law major. Do you have any good suggestions?)\\
        & 我准备学酒店管理,我喜欢甜品 (I'm going to learn Hotel Management. I like dessert) \\
        & 我想报计算机,你有什么好的建议吗? (I want to study Computer Science. Do you have any good suggestions?)\\
        & 我啊,最看重学校,其次是专业。 (For me, school is the most important, followed by major.)\\
        \midrule
       \multirow{5}{*}{Base:} & 我要学医,你呢? (I want to study Medicine. How about you?)\\
        & 我是学医的,以后可能会做医生 (I am a medical student. I may become a doctor in the futur)\\
        & 我的专业是应用化学,你呢? (My major is Applied Chemistry. What about you?)\\
        & 我想报计算机,你觉得怎么样? (I want to study Computer Science, What do you think?)\\
        & 我要学医,你呢? (I want to study Medicine. How about you?)\\
        \midrule
        \textit{context:}    & 这个周末我们公司要搞团建！(This weekend, our company will carry out league construction!)\\
        \midrule
        \multirow{5}{*}{DDS:}& 这个好,我们学校也搞过 (This is good. Our school has also done it)\\
        & 这个主题不错 (This theme is good)\\
        & 我们也做过哎 (We did it too)\\
        & 我们也做过 (We did it too)\\
        & 这个主意不错 (This is a good idea)\\
        \midrule
        \multirow{5}{*}{Base:} & 啊,你们团建是什么 (Ah, what are you going to do in league construction)\\
        & 这个主意不错(This is a good idea)\\
        & 我知道啊,但是我不去,你组织 (I know, but I won't go, you organize)\\
        & 是我们这边的郊区的团建吗 (Is it the group building in the suburb of our side)\\
        & 我们没有这个活动吧 (We don't have this activity, do we?)\\
    \bottomrule
    \end{tabular}
    \caption{Generated samples from Diamante.}
    \label{tb:my_label}
\end{table}

\begin{table}[!ht]
    \centering
    \small
    \begin{tabular}{r|p{5.2cm}}
    \toprule
      \textit{context:}    &  谢婉莹是谁？ (Who is Xie Wanying?)\\
      \textit{true answer:} & 中国诗人，现代作家、翻译家、儿童文学作家、社会活动家 (Chinese poet, modern writer, translator, children's literature writer, social activist)\\
      \midrule
       \multirow{5}{*}{DDS:} & 谢婉莹2012年农历腊月22日创作并发表了单曲《我不是你爸爸的爸爸》 (Xie Wanying wrote and published the single "I'm not your father's father" on the 22nd lunar month of 2012)\\
        & 女,中国诗人,现代作家、翻译家 (Female, Chinese poet, modern writer and translator)\\
        & 中国古典小说《甄嬛传》中的主要角色之一 (One of the main characters in the classical Chinese novel Legend of Zhen Huan) \\
        & 中国诗人,现代作家、翻译家、儿童文学作家、社会活动家 (Chinese poet, modern writer, translator, children's literature writer, social activist)\\
        & 中国诗人,现代作家、翻译家、儿童文学作家 (Chinese poet, modern writer, translator, children's literature writer)\\
        \midrule
       \multirow{5}{*}{Base:} & 电视剧《永不言弃》中的角色 (The role in the TV play Never Give Up)\\
        & 小说《蜀山剑侠传》的女主人公 (The heroine of the novel "Legend of the Swordsman of Shu Mountain")\\
        & 中国诗人,现代作家。著有诗集《黄昏》 (Chinese poet and modern writer. He is the author of the poetry collection "Twilight")\\
        & 现代作家。代表作有《红高粱家族》 (Modern writers. The representative work is Red Sorghum Family)\\
        & 女,教授,南京师范大学商学院副教授 (Female, professor, associate professor of Nanjing Normal University Business School)\\
        \midrule
        \textit{context:}    & 熊黛林多高？(How tall is Xiong Dailin?)\\
        \textit{true answer:} & 178cm\\
        \midrule
        \multirow{5}{*}{DDS:}& 178cm \\
        & 178cm \\
        & 178cm \\
        & 178cm\\
        & 178cm \\
        \midrule
        \multirow{5}{*}{Base:} & 168cm \\
        & 168cm \\
        & 178cm \\
        & 178cm \\
        & 178cm \\
    \bottomrule
    \end{tabular}
    \caption{Generated samples from PersonQA.}
    \label{tb:my_label2}
\end{table}

\section{Details about QA Datasets}
\label{QA_data}
Both PersonQA and LQA are our internal data, and each sample is a question-answer pair. Questions are collected from Internet, and answers are manually annotated. Both questions and answers will be classified. PersonQA contains QA pairs about persons and LQA includes samples that have longer explanations. All the data is anonymized and nonsensitive.

\section{Insights for Mapping Strategies}
We design three mapping strategies in order to cover all major types of mapping trends. Specifically, As shown in Figure~\ref{fig:mapping}, Linear mapping simply projects the diversity score to temperature linearly. Exponential mapping has flat slope when diversity score is near the mean value while sharp slope at either end. Conversely, Inverse Sigmoid mapping shows a different trend. According to Section~\ref{sc:map}, all three types of strategies can work on both two scenarios.